\documentclass[conference]{IEEEtran}
\IEEEoverridecommandlockouts
\usepackage{cite}
\usepackage{amsmath,amssymb,amsfonts}
\usepackage{algorithmic}
\usepackage{graphicx}
\usepackage{textcomp}
\usepackage{xcolor}
\usepackage{array,multirow}
\usepackage{diagbox}
\usepackage{slashbox}
\usepackage{booktabs}
\def\BibTeX{{\rm B\kern-.05em{\sc i\kern-.025em b}\kern-.08em
    T\kern-.1667em\lower.7ex\hbox{E}\kern-.125emX}}
    
\begin{document}



\title{Instance Retrieval at Fine-grained Level Using Multi-Attribute Recognition}

\author{\IEEEauthorblockN{Roshanak Zakizadeh\IEEEauthorrefmark{1}, Yu Qian\IEEEauthorrefmark{1}, Michele Sasdelli\IEEEauthorrefmark{2} and Eduard Vazquez\IEEEauthorrefmark{1}}
\IEEEauthorblockA{\IEEEauthorrefmark{1}\textit{Cortexica Vision Systems Limited}, London, UK \\}
\IEEEauthorblockA{\IEEEauthorrefmark{2}\textit{Australian Institute for Machine Learning, the University of Adelaide, Australia}}
Email: roshanak.zakizadeh@cortexica.com}


\maketitle

\begin{abstract}
In this paper, we present a method for instance ranking and retrieval at fine-grained  level based on the global features extracted from a multi-attribute recognition model which is not dependent on landmarks information or part-based annotations. Further, we make this architecture suitable for mobile-device application by adopting the bilinear CNN to make the multi-attribute recognition model smaller (in terms of the number of parameters). The experiments run on  the Dress category of DeepFashion In-Shop Clothes Retrieval and CUB200 datasets show that the results of instance retrieval at fine-grained level are promising for these datasets, specially in terms of texture and color.
\end{abstract}


\section{Introduction}
Content-based image retrieval (CBIR) is the task of identifying relevant images using the representative visual content (such as high-level information in an image)~\cite{smeulders2000content,lew2006content,liu2007survey,zhou2017recent}. On the other hand, instance retrieval at fine-grained level can be defined as a finer visual search, for example finding a dress with a design or pattern similar to the query's from the catalog of dresses or retrieving a certain species of bird. Fine details (namely attributes) help to tell apart different instances with similar appearances, such as two similar gulls of different types. This work discusses instance retrieval at fine-grained level which we briefly call instance retrieval through the rest of this paper to differentiate from CBIR.  

\begin{figure}
    
     \centerline{\includegraphics[trim={0.2cm 4cm 0.2cm 2cm},clip, width=\linewidth]{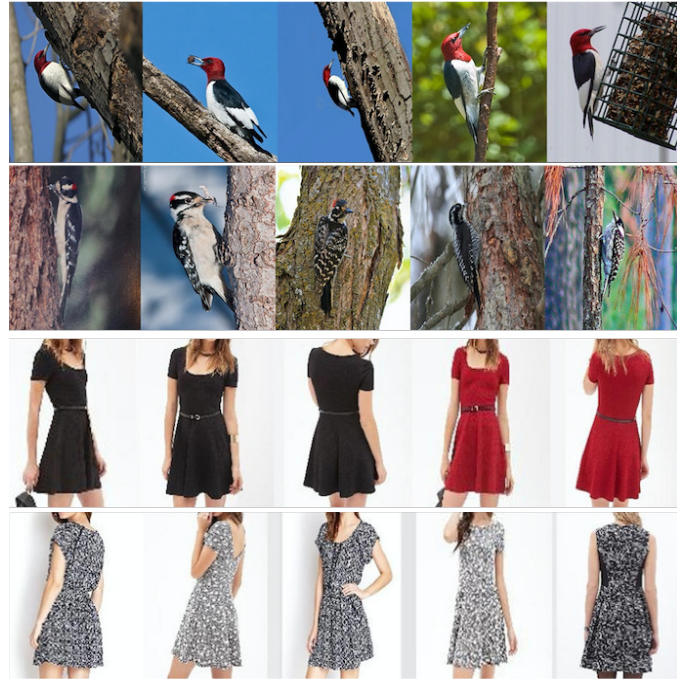}}
    \caption{First image from left in all rows are the query images. Rows one and three: Examples of correct instance retrieval at fine-grained level where the exact item has been retrieved using the proposes method (row 1: retrieving red headed woodpecker from CUB200~\cite{wah2011caltech} and dress item 388 from Deepfashion In-shop Retrieval dataset~\cite{liu2016deepfashion}). Rows two and four: Examples of where the close items are retrieved but not always the exact same item (Row 2: we can see that four instances which are retrieved are woodpecker but not all four are Downy woodpecker. Row 4: In this case only the second retrieved dress (third image from the left) is an exact match, nonetheless, the other three retrieved items are extremely similar in terms of texture.)}
    \label{fig:teaser}
\end{figure}

The importance of semantic attributes in instance retrieval has been emphasized before~\cite{siddiquie2011image}. Most of the research in fine-grained instance retrieval use the attributes directly to retrieve instances~\cite{siddiquie2011image,rastegari2013multi,cao2014image}. Another alternative is to use the features extracted from a pre-trained network on Imagenet~\cite{wan2014deep,yue2015exploiting,zhou2017recent,razavian2016visual} (similar to some approaches in CBIR). However, this is not a very accurate approach for instance retrieval, since Imagenet~\cite{russakovsky2015imagenet} includes 1000 object classes which are mostly categorized at a coarse level of recognition, for instance there are a few categories of birds but not as fine-grained as a dataset like CUB200~\cite{wah2011caltech}. On the other hand, fine-tuning a pre-trained model on Imagenet for a fine-grained dataset requires the prior knowledge of classes in that dataset. However, in fine-grained instance retrieval, very often we receive datasets of images with no specific categories or classes and the only available information is the annotated (or extracted from the meta data) attributes. This makes fine-grained instance retrieval different from fine-grained instance classification~\cite{lin2015bilinear,zhang2014part}; a problem which is often addressed with regards to fine-grained visual recognition. 
In this work, we show that it is possible to achieve good instance retrieval results using the global features extracted from a trained multi-attribute recognition network. Similar to CBIR, the global features (e.g. from a fully connected layer) are used by a metric learning method (e.g. euclidean distance) to retrieve similar instances to the query image; however, here the network is trained for attribute recognition at fine-grained level. We choose VGG16 architecture~\cite{simonyan2014very} as it has small filter kernels which make capturing small details possible. To adopt VGG16 for multi-attribute recognition  we use pair-wise ranking loss function~\cite{li2017improving} which is proved to be efficient for multi-label classification. This simple approach leads to competitive results and it is able to capture similar instances, specially with a good visual similarity in terms of texture, color, material and shape. Further, by adopting Bilinear CNN~\cite{lin2015bilinear} we modify the VGG16~\cite{simonyan2014very} network and introduce a small model (in terms of number of parameters) for multi-attribute recognition at fine-grained level, which can achieve satisfactory results. The size of the network is always an important factor to consider at the production level.

We are experimenting with two datasets: CUB200 dataset~\cite{wah2011caltech} which consists of 11k images of 200 species of birds, annotated with 312 attributes and dress category from DeepFashion In-shop Retrieval dataset~\cite{liu2016deepfashion} with 336 attributes. Previous research in clothes retrieval has rarely considered a fine-grained case where only one category of clothes (e.g. dress) is available to train on and the retrieval task has to be done against a gallery of diverse clothing types. We prove it is possible to achieve good retrieval results when the network is only exposed to images of one category, similar to that of the query image (e.g. when a customer provides a catalogue of dresses for which the attributes are known or can be extracted from the metadata and he/she is interested in finding similar items to the ones in the catalogue but from a diverse dataset of clothes.). Further, most of the clothes retrieval techniques often heavily rely  on landmark detection~\cite{liu2016deepfashion}, whereas, in our approach we are ignoring landmark information.

The paper is organized as follows. In Section~\ref{sec:data}, we explain the datasets used in the experiments. In Section~\ref{sec:method}, we first discuss the network used for multi-attribute recognition and the smaller network which is based on the bilinear CNN architecture. Then we explain what global features are used for retrieval and eventually, which metric learning methods are used for retrieving a query item from the gallery.  The results are presented in Section~\ref{sec:results} and the paper is finally concluded in Section~\ref{sec:conclude}.

\section{Datasets}
\label{sec:data}
There are a few fine-grained datasets publicly available~\cite{maji2013fine,wah2011caltech,yang2015large}, among which some provide annotated attributes for fine-grained parts in addition to the instance classes~\cite{maji2013fine,wah2011caltech} and some do not provide any attribute annotations~\cite{yang2015large}. Here, we have chosen CUB200~\cite{wah2011caltech} which includes 11k images of 200 birds species. The species are classified at fine-grained level. An example can be seen in Fig.~\ref{fig:teaser} where all instances in the first row are the same type of woodpecker, whereas, in the second row we can see different species of woodpeckers (the differences are very subtle even for human observers). The birds are annotated with 312 attributes including the colors for different parts, beak shape, etc. The problem with the CUB200 annotations is that the list of the annotated attributes per item is long and not very distinctive which makes the task of retrieval based on attributes more difficult.

DeepFashion In-shop Retrieval dataset~\cite{liu2016deepfashion} consists of several categories of clothes for men and women and overall 465 attributes. The dataset is designed specifically for retrieval, therefore, for each query image there are similar items (some in different colors) available in the gallery. Deepfashion is not a fine-grained dataset, however, each category of clothes can make a fine-grained case. We have chosen the dress category which is annotated with 336 attributes. The dataset also provides the bounding box for each item as well as landmark points. In our experiments, we are not using the landmark information and only crop the images within the given bounding box to lessen the effect of faces on the network training process. We are experimenting with both the gallery of dresses only and the gallery of all categories to compare the latter's results with the benchmark results by FashionNet~\cite{liu2016deepfashion} for the dress category of In-shop  Retrieval dataset.



\section{Retrieval Method}
\label{sec:method}
\subsection{Multi-Attribute Recognition Network}
\begin{figure*}[!h]    
     \centerline{\includegraphics[trim={1cm 1cm 3cm 6cm},clip, height=8cm, width=\linewidth]{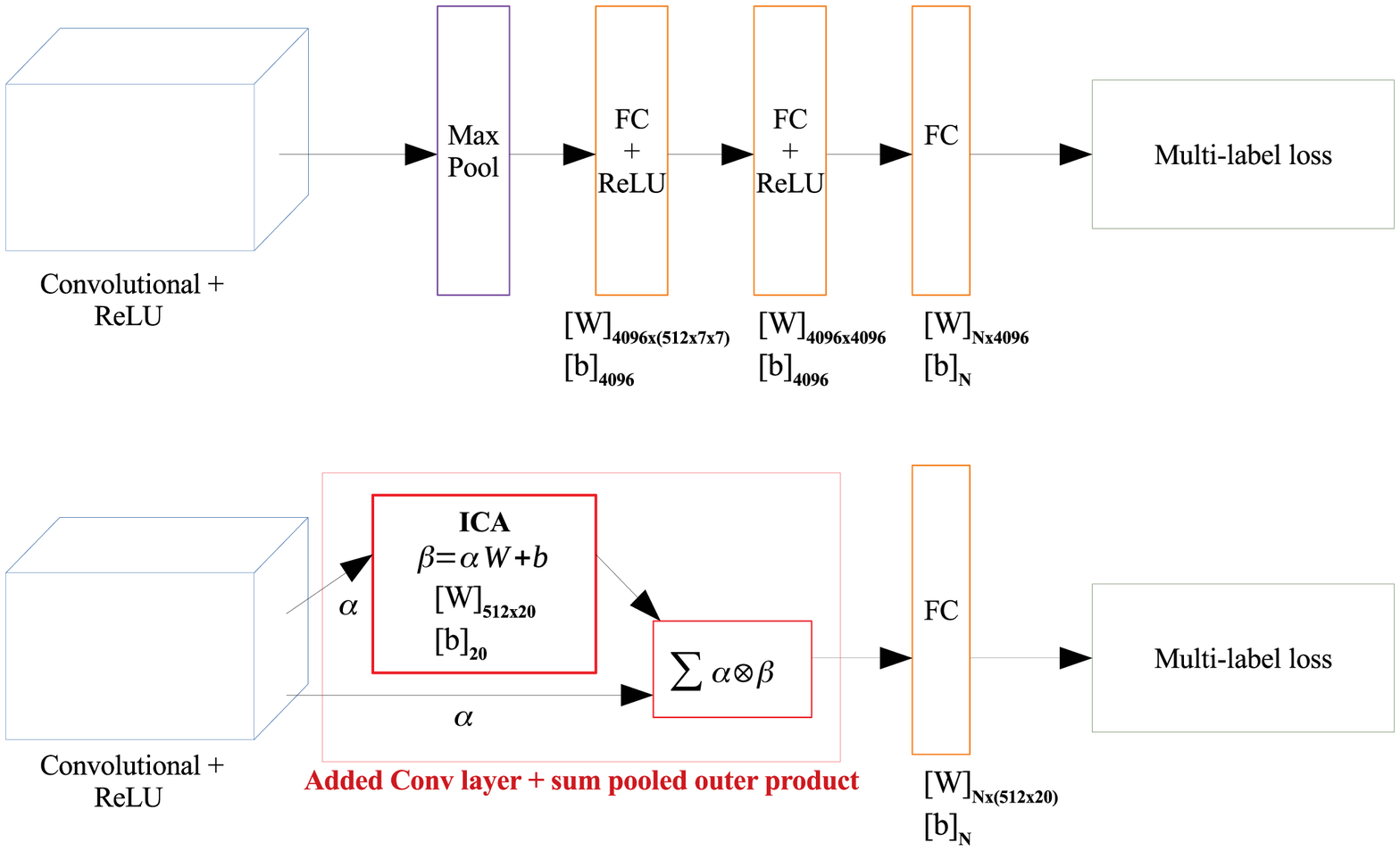}}
    \caption{The VGG16 based multi-attribute recognition networks with pairwise ranking loss for attribute extraction at fine-grained level. In the bottom,  the bilinear CNN architecture is shown.}
    \label{fig:net}
\end{figure*}

For multi-attribute recognition at fine-grained level we are using VGG16 architecture~\cite{simonyan2014very} (the top network in Fig.~\ref{fig:net}) which is a convolutional neural network with small convolutional filter kernels ($3\times3$) which makes it suitable for capturing fine details of textures in an image. 

We have adopted VGG16 for multi-attribute recognition by using a smooth pairwise ranking loss function ~\cite{li2017improving}:

\begin{equation}
 L_{log\_sum\_exp} = log\left( 1 + \sum_{v\notin Y, u \in Y} \exp \left ( f_v(x) - f_u(x) \right ) \right ),  
\label{eq:smooth_rank_loss}
\end{equation}

where $f(x) : \mathbb{R}^d \rightarrow \mathbb{R}^K$ is a label (attribute) prediction model that maps an image to a K-dimensional label space which represents the confidence scores. The model $f(x)$ is designed such that it produces a vector whose values for true labels are greater than those for negative labels (i.e. $f_u(x)>f_v(x),$ $\forall u \in Y, v \notin Y$). The loss function in (\ref{eq:smooth_rank_loss}) enforces this property by calculating the log-sum-exp of all pairs of labels (attributes) and penalizing the values which do not follow the mentioned rule. This creates the framework of learning to rank~\cite{liu2009learning} via pairwise comparisons. Equation (\ref{eq:smooth_rank_loss}) is a smooth approximation of a similar hinge function~\cite{gong2013deep,weston2011wsabie} used for pairwise comparison. The smooth version proposed by~\cite{li2017improving} makes optimization easier due it its differentiability.

The second architecture is also based on VGG16; however, to capture more details we adopt the bilinear CNN architecture~\cite{lin2015bilinear} and place a bilinear layer (shown in the red box in Fig.~\ref{fig:net}) right after the last convolutional layer (Conv5\_3 + Relu). To save the space and reduce the correlation between the feature maps (hence, capturing more details) the second copy of the feature map ($\beta$ in Fig.~\ref{fig:net}) is generated by projecting a copy of the original feature map ($\alpha$) into a 20 dimensional  ICA projection space~\cite{hyvarinen1999survey} (which is generated beforehand based on the feature maps from the same training set). Then, the sum of the outer product of $\alpha$ and $\beta$ at each location is calculated which is passed through a fully connected layer.  The same smooth pairwise ranking loss (\ref{eq:smooth_rank_loss}) is used to learn the ranked list of attributes.

In both networks, the number of labels ($N$ in Fig.~\ref{fig:net}) equals the number of attributes describing the dataset which is 312 for CUB200 dataset and  336 for the dress category in DeepFashion In-shop Retrieval dataset. The important advantage of the second architecture is the size of this network in terms of parameters. For a vocabulary of roughly 300 words (similar to the one for CUB200 or dress category form the Deepfashion) the second architecture is $40\times$ smaller, which makes it  suitable for mobile-device application. The number of parameters for the fully connected layers in VGG16 is shown in Fig.~\ref{fig:net}, we can see that two of these layers are replaced by the bilinear layer which uses a projection space to reduce the dimension of one copy of Conv5\_3 + Relu feature map which results in a great overall save of space.

For the DeepFashion In-shop retrieval dataset, our models are trained \textbf{only} on the \textbf{dress category} and not all the categories, since we are interested in the case where only same category of clothes as the query image is available for training. A practical scenario is when a customer provides a catalogue of one type of clothes (e.g. dresses) and is interested in finding dresses with the same pattern or design within a set of diverse clothes.  

To train the models, VGG16 pre-trained weights~\cite{deng2009imagenet,krizhevsky2012imagenet} on the Imagenet dataset~\cite{russakovsky2015imagenet} are used for initializing the convolutional layers in the second network and for all layers in the first architecture.

The models are built on Tensorflow framework and the experiments are run on an NVIDIA Tesla V100 GPU. For each category the model is trained in average for 14 epochs with the batch size of 16 using Adam optimizer~\cite{kingma2014adam} with the base learning rate of 0.00001. 

Through the rest of the paper we call the first architecture ``VGG16 + MultiAttrib" and the second one ``bilin + MultiAttrib".


\subsection{Features Used for Retrieval}
We are using global features from both networks.  We have experimented with the last three fully connected layers: fc6,fc7 and fc8 from VGG16 + MultiAttrib network, out of which fc6 features resulted in better retrieval results compared to the other two. From bilin + MultiAttrib architecture, the feature map from the bilinear layer are used as well as the outputs of the network, i.e. the scores given to $n$ attributes  (312 attributes of the CUB200 dataset and 336 attributes of the  dress dataset), this is annotated as `prob' in Table~\ref{tab:results}. 

As mentioned before, it is common to use the features of a pre-trained network on Imagenet dataset for content-based image retrieval. Here, we are comparing our results with the ones from the fc6 layer of the pre-trained VGG16 on imagenet dataset (again, we experimented with different layers and found fc6 features to be slightly better).


\subsection{Metric Learning}
For most of the experiments the global features are L2 normalized and then the Euclidean distance is used for retrieving the query from the set of images. However, we found out that histogram intersection works better when using prob features (the scores) from the bilin + MultiAttrib network for retrieval.


\section{Experiments and Results}
\label{sec:results}
As mentioned before three global feature maps are used for instance retrieval in our experiments: 1- the fc6 layer of VGG16 + MultiAttrib network, 2- the bilinear layer from bilin + MultiAttrib network and 3- the scores from bilin + MultiAttrib network (prob). 

Table~\ref{tab:results} shows the retrieval results for dress category of DeepFashion In-shop Retrieval dataset (the top section of the table) and CUB200 dataset (the bottom section of the table). The model used for producing the features are mentioned in the first column of the table. Second column shows which global features are used for retrieval and fourth column lists the metric learning methods. We also show the number of features for each case in the third column of table. The bilinear layer feature map (from bilin + MultiAttrib network) has 10240 features which is the size of the outer product of the original 512 features from the previous layer (Conv5\_3 + Relu) multiplied by the size of the reduced copy of it (which is 20). Also, the size of prob (scores by bilin + MultiAttrib network) equals the number of attributes for each dataset.

\begin{table*}[ht]
\caption{Instance Retrieval precision and attribute similarity results for Deepfashion In-shop Retrieval Dress and CUB200 dataset.}
\centering
\begin{tabular}{|@{}cccclllllllll@{}|}
\hline
\multicolumn{13}{|c|}{Deep-fashion In-shop Retrieval Dress}                                                                                                                                 \\ \hline
                                  & &    & \multicolumn{1}{c|}{}         & \multicolumn{6}{@{}c@{}|}{Fine-level Similarity (mAP)}     & \multicolumn{3}{@{}c@{}|}{Attribute Similarity (IoU)} \\ \hline
                                  
\multicolumn{1}{|@{}c|}{model}            & \multicolumn{1}{c|}{layer}    & \multicolumn{1}{c|}{feature size}  & \multicolumn{1}{c|}{metric}     &  \multicolumn{2}{c}{top-1} &  \multicolumn{2}{c}{top-5} & \multicolumn{2}{c|}{top-10}  & top-1     & top-5     & top-10    \\ \hline

\multicolumn{1}{|@{}c|}{VGG16 (imagenet)} & \multicolumn{1}{c|}{fc6}     &  \multicolumn{1}{c|}{4096} & \multicolumn{1}{c|}{L2 + Euclidean} &     \multicolumn{2}{c}{0.33} &   \multicolumn{2}{c}{0.15}   &  \multicolumn{2}{c|}{0.10} &   0.48  &   0.35 &   0.31 \\
\multicolumn{1}{|@{}c|}{VGG16 + MultiAttrib}       & \multicolumn{1}{c|}{fc6}     & \multicolumn{1}{c|}{4096} & \multicolumn{1}{c|}{L2 + Euclidean} &      \multicolumn{2}{c}{\textbf{0.74}} &   \multicolumn{2}{c}{0.36}   &  \multicolumn{2}{c|}{0.22} &  \textbf{0.81}  &  0.52 &  0.42 \\
\multicolumn{1}{|@{}c|}{bilin + MultiAttrib}       & \multicolumn{1}{c|}{bilinear}   &  \multicolumn{1}{c|}{10240} & \multicolumn{1}{c|}{L2 + Euclidean} &     \multicolumn{2}{c}{0.70} &   \multicolumn{2}{c}{0.37}   &  \multicolumn{2}{c|}{0.23}    &  0.77 &  0.53   &   0.42     \\

\multicolumn{1}{|@{}c|}{bilin + MultiAttrib}       & \multicolumn{1}{c|}{prob}       &  \multicolumn{1}{c|}{336} & \multicolumn{1}{c|}{L2 + Hist\_inter} &     \multicolumn{2}{c}{0.69} &   \multicolumn{2}{c}{\textbf{0.39}}   &  \multicolumn{2}{c|}{\textbf{0.24}}    &      0.77    &     \textbf{0.55}     &    \textbf{0.45}     \\
\hline
\multicolumn{13}{|c|}{CUB200}                                                                                                                                                               \\ \hline

                  & & & \multicolumn{1}{c|}{}         & \multicolumn{3}{@{}c@{}|}{Fine-level Similarity (mAP)}         & \multicolumn{3}{@{}c@{}|}{Coarse-level Similarity (mAP)}       & \multicolumn{3}{@{}c@{}|}{Attribute Similarity (IoU)} \\ \hline
\multicolumn{1}{|@{}c|}{model}            & \multicolumn{1}{c|}{layer}    & \multicolumn{1}{c|}{feature size}  & \multicolumn{1}{c|}{metric} & top-1 & top-5 & \multicolumn{1}{l|}{top-10} & top-1 & top-5 & \multicolumn{1}{l|}{top-10} & top-1     & top-5     & top-10    \\ \hline

\multicolumn{1}{|@{}c|}{VGG16 (imagenet)} & \multicolumn{1}{c|}{fc6}  &  \multicolumn{1}{c|}{4096} & \multicolumn{1}{c|}{L2 + Euclidean}    & 0.40   & 0.31   & \multicolumn{1}{l|}{0.27}   & 0.59   & 0.53   & \multicolumn{1}{l|}{0.48}   & 0.27       & 0.27       & 0.27 \\

\multicolumn{1}{|@{}c|}{VGG16 + MultiAttrib}       & \multicolumn{1}{c|}{fc6}  & \multicolumn{1}{c|}{4096} & \multicolumn{1}{c|}{L2 + Euclidean}    & 0.41   & 0.32   & \multicolumn{1}{l|}{0.28}   & 0.52   & 0.53   & \multicolumn{1}{l|}{0.50}   & 0.28       & 0.29       & 0.28      \\

\multicolumn{1}{|@{}c|}{bilin + MultiAttrib}       & \multicolumn{1}{c|}{bilinear} &   \multicolumn{1}{c|}{10240} & \multicolumn{1}{c|}{L2 + Euclidean}    &    \textbf{0.43}  & \textbf{0.37} & \multicolumn{1}{l|}{\textbf{0.33}} &  \textbf{0.63}    &  \textbf{0.58}    & \multicolumn{1}{l|}{\textbf{0.54}}     &  0.29   &  0.29        & 0.29        \\

\multicolumn{1}{|@{}c|}{bilin + MultiAttrib}       & \multicolumn{1}{c|}{prob}      &   \multicolumn{1}{c|}{312} & \multicolumn{1}{c|}{L2 + Hist\_inter}   & 0.42 & 0.30   & \multicolumn{1}{l|}{0.26}  &  0.55  & 0.50     & \multicolumn{1}{l|}{0.47}     &  0.29   &   0.29   &  0.29  \\ \hline
\end{tabular}
\label{tab:results}
\end{table*}

For the dress dataset the query is retrieved from the gallery of dresses only. The first three columns of the results are the mean average precision~\cite{schutze2008introduction} of top-1, top-5 and top-10 retrieval calculated as the ratio between the relevant retrieved items and k retrieved items, for instance if in top-5 retrieval results 3 items are retrieved correctly the precision is calculated as 0.60. The reported results are the mean average precision over the whole 1901 query images of dresses. The last three columns in the table show the attribute similarity precision which is calculated as the Intersection over Union between the attributes of the query image and the retrieved items from the gallery. We can see that the best top-1 precision results belong to the fc6 features of VGG16 + MultiAttrib network. In top-5 and top-10 retrieval results the output (prob) from the bilin + MultiAttrib network outperforms the other two global features. The first row shows the global features from the fc6 layer of the pre-trained VGG16 on Imagenet. We can see that using the global features from the pre-trained VGG16 network on Imagenet for instance retrieval leads to poor results compared to the ones by global features of the multi-attribute recognition network.

The bottom section of the table shows the same results for CUB200 dataset. In addition to the fine-grained retrieval results, we are also reporting the results for the retrieval precision at coarse level. An example of which can be seen in the second row of Fig.~\ref{fig:teaser} where all retrieved items are woodpeckers but not all are Downy woodpecker (which is the query item). For CUB200 dataset we can see that the best results are achieved using the feature maps from the bilinear layer of bilin + MultiAttrib network.

Another point to notice is that the performance for birds is not as good as dress dataset since the attribute annotation for birds in general is poorer than dresses, i.e. there is a lot of overlap between the attributes for different species and the list of attributes is very long and less distinctive.

\begin{figure}[ht]
     \centerline{\includegraphics[trim={1.1cm 0.3cm 0.5cm 0.4cm},clip,width=\linewidth]{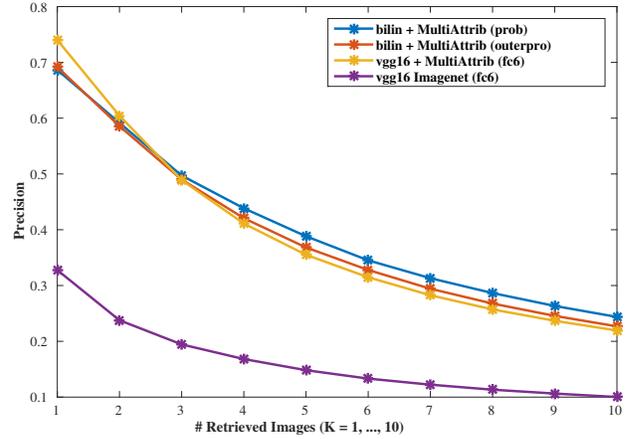}}
    \caption{The precision of top-k retrieval results for the two global feature maps (prob and bilinear (annotated as outerpro)) of the bilin + MultiAttrib network and the features of fc6 layer from VGG16 + MultiAttrib  network for the dress category of DeepFashion In-shop Retrieval dataset. The performance of the fc6 layer features extracted from the pre-trained VGG16 on Imagenet  is also shown.}
    \label{fig:dress_gallery}
\end{figure}

More detailed precision results (for all top-k retrieval k=1, ..., 10) for the dress dataset are shown in Fig.~\ref{fig:dress_gallery}.

\begin{figure}[ht]
     \centerline{\includegraphics[trim={1.1cm 0.3cm 0.5cm 0.4cm},clip,width=\linewidth]{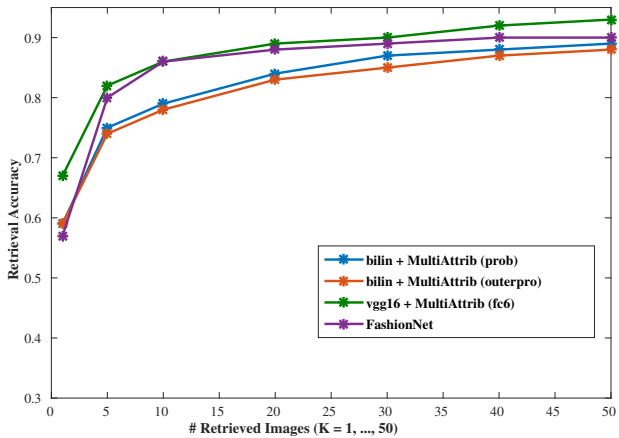}}
    \caption{Top-k retrieval accuracy results for retrieving the query dress from the gallery of all categories of clothes for the two global feature maps (prob and bilinear (annotated as outerpro)) of the bilin + MultiAttrib network and the features of fc6 layer from VGG16 + MultiAttrib  network. The results are compared with the FashionNet~\cite{liu2016deepfashion} results for the dress category.}
    \label{fig:dress_allgallery}
\end{figure}

To compare the retrieval results for the In-shop Retrieval dress dataset with the state of the art results we have retrieved the query dresses against the whole gallery consisting of clothes from all categories (including dresses) and compared the results with the ones by FashionNet~\cite{liu2016deepfashion}. The results are plotted in Fig.~\ref{fig:dress_allgallery} for top-k retrieval (k=1, ..., 50). We are using the same evaluation technique proposed by FashionNet authors who calculate retrieval accuracy based on successful retrievals which is defined as finding at least one similar item to the query in top-k results. We can see that the top-1 retrieval results for all three global feature maps extracted from our multi-attribute recognition networks outperform the FashionNet results. Further, we can see that fc6 features from VGG16 + MultiAttrib network is always leading for all top-k retrieval results. It is important to notice that we are comparing a much simpler technique with a complicated network such as FashionNet which makes use of landmark information and the retrieval results are also learned by the triplet loss. The simplicity of our proposed method makes it more suitable for practical applications. Further, we need to consider the fact that the multi-attribute recognition networks used in our experiment have never been exposed to other categories of clothes and are only trained on the dress training set. This addresses the scenario where only the catalogue of one type of clothes desired by a costumer is provided for training and we need to query against a diverse clothes dataset.

\begin{figure*}[ht]  
     \centerline{\includegraphics[trim={5cm 0cm 5cm 0cm},clip, height = 18cm,width=0.8\linewidth]{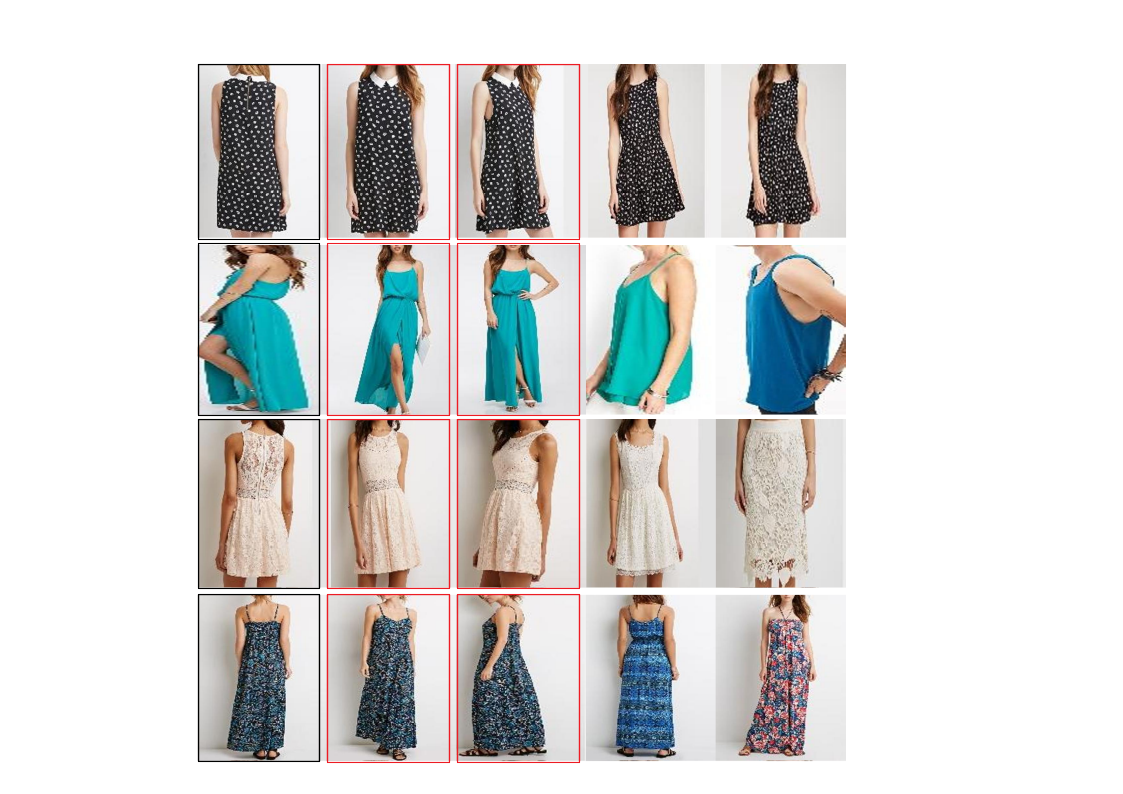}}
    \caption{Performance of fc6 layer global features from the VGG16 architecture adopted for multi-attribute recognition (with pairwise ranking loss) which is trained on dress category of Deepfashion In-shop Retrieval dataset. Row 1: example of similar texture retrieval (dotted), row 2: example of similar color retrieval (green/blue), row 3: example of similar fabric retrieval (lace) and row 4: example of similar style retrieval (maxi strapped dress) (the exact retrieved item is shown with a red rectangle).}
    \label{fig:differentmodules}
\end{figure*}

Fig.~\ref{fig:differentmodules} demonstrates the visual examples of the retrieval results using fc6 features from VGG16 + MultiAttrib network for different types of attributes. The query item is marked in a black rectangular box and the correctly retrieved items are shown in red rectangles. The first row is an example of successful retrieval of texture by our method where the first two retrieved items are exactly the same as query and the last two dresses are very similar in terms of texture. The second row shows that all the retrieved items are from the same hue (green/bluish green). Fabric is another important factor in clothes retrieval, an example of the successful retrieval of similar fabrics by our method can be seen in the third row where all the retrieved dresses are made of lace. Finally, the last row of Fig.~\ref{fig:differentmodules} shows that the retrieved dresses are all maxi with strapped shoulders which is an example of successful retrieved styles by fc6 features of VGG16 + MultiAttrib.

\section{Conclusions}
\label{sec:conclude}
In this paper, we showed that by using the global features from the multi-attribute recognition network we can achieve successful instance retrieval results at a fine-grained level. We concluded that for both CUB200 and DeepFashion IN-Shop Retrieval dress datasets the instance retrieval results using the global features of the multi-attribute recognition networks are better than the ones by the global features from the pre-trained network on Imagenet. We demonstrated that for the dress category of DeepFashion In-shop Retrieval dataset we can get competing retrieval results in comparison to the benchmark FashionNet method. These results are significant considering the fact that our proposed method is oblivious to the landmark information and it is simpler to implement. Besides, it addresses the scenario where only one category of clothes with annotated attributes is provided for training, but the retrieval needs to be done from a diverse set of clothes. Further, we showed that by adopting bilinear CNN architecture we can reduce the size of the network to $40\times$ smaller than the original VGG16 and still achieve good retrieval results using global features extracted from the model. The latter design makes the model suitable for mobile-device application. The visual analysis of the retrieved results for the dresses confirms the efficiency of our method in retrieving similar items in terms of texture, color, fabric and design.

\bibliographystyle{IEEEtran}
\bibliography{ref.bib}
\vspace{12pt}

\end{document}